\documentclass[lettersize,journal]{IEEEtran}
\usepackage{amsmath,amsfonts}
\usepackage{array}
\usepackage[caption=false,font=footnotesize,labelfont=rm,textfont=rm]{subfig}
\usepackage{textcomp}
\usepackage{stfloats}
\usepackage{url}
\usepackage{verbatim}
\usepackage{graphicx}
\usepackage{cite}
\usepackage{graphicx}
\usepackage{booktabs}

\usepackage{xcolor}
\ifodd 0

\else
\fi
\usepackage{multirow}

\usepackage[ruled,vlined]{algorithm2e}
\usepackage{amsmath, amssymb}
\usepackage{setspace}

\begin{document}

\title{SL-FAC: A Communication-Efficient Split Learning Framework with Frequency-Aware Compression}

\author{Zehang Lin,  Miao Yang, Haihan Zhu, Zheng Lin, Jianhao Huang, Jing Yang, Guangjin Pan, Dianxin Luan, Zihan Fang, Shunzhi Zhu, Wei Ni,~\IEEEmembership{Fellow,~IEEE},
and John Thompson,~\IEEEmembership{Fellow,~IEEE}
\thanks{Z. Lin, M. Yang, X. Du, and S. Zhu are with the School of Computer and Information
 Engineering, Xiamen University of Technology, Xiamen, 361000, China (email: 18850975878@163.com; yangmiao@xmut.edu.cn; duxia@xmut.edu.cn; szzhu@xmut.edu.cn). }
 \thanks{H. Zhu is with the School of Artificial Intelligence, Hebei University of Technology (email: 242794@stu.hebut.edu.cn)}
 \thanks{Z. Lin and J. Huang are with the Department of Electrical and Electronic Engineering, The University of Hong Kong, Hong Kong. (e-mail: linzheng@eee.hku.hk; jianhaoh@hku.hk)}
 \thanks{J. Yang is with the Center of Research for Cyber
Security and Network (CSNET), Faculty of Computer Science and Information Technology, Universiti Malaya, 50603 Kuala Lumpur, Malaysia (e-mail:
s2147529@siswa.um.edu.my).}
 \thanks{G. Pan is with the Department of Electrical Engineering, Chalmers University
of Technology, 41296 Gothenburg, Sweden (e-mail: guangjin.pan@chalmers.se)}
\thanks{D. Luan and J. Thompson are with the Institute for Imaging, Data and Communications,
School of Engineering, The University of Edinburgh, EH9 3JL Edinburgh, U.K. (e-mail: dianxin.luan@ed.ac.uk; john.thompson@ed.ac.uk).}
\thanks{Z. Fang is with the Department of Computer Science, City University
of Hong Kong, Kowloon, Hong Kong SAR, China (e-mail: zihanfang3-
c@my.cityu.edu.hk).}
\thanks{W. Ni is with the School of Engineering, Edith Cowan University, Perth, WA 6027, Australia (email: wei/ni@ieee.org).}
}

\markboth{}%
{Shell \MakeLowercase{\textit{et al.}}: A Sample Article Using IEEEtran.cls for IEEE Journals}

\maketitle

\begin{abstract}
The growing complexity of neural networks hinders the deployment of distributed machine learning on resource-constrained devices. Split learning (SL) offers a promising solution by partitioning the large model and offloading the primary training workload from edge devices to an edge server. However, the increasing number of participating devices and model complexity leads to significant communication overhead from the transmission of smashed data (e.g., activations and gradients), which constitutes a critical bottleneck for SL. To tackle this challenge, we propose SL-FAC, a communication-efficient SL framework comprising two key components: adaptive frequency decomposition (AFD) and frequency-based quantization compression (FQC). AFD first transforms the smashed data into the frequency domain and decomposes it into spectral components with distinct information. FQC then applies customized quantization bit widths to each component based on its spectral energy distribution. This collaborative approach enables SL-FAC to achieve significant communication reduction while strategically preserving the information most crucial for model convergence. Extensive experiments confirm the superior performance of SL-FAC for improving the training efficiency.
\end{abstract}

\begin{IEEEkeywords}
Distributed learning, split learning, communication efficiency, and adaptive frequency decomposition.
\end{IEEEkeywords}

\section{Introduction}
\label{sec:intro}

The proliferation of Internet of Things (IoT) devices is generating massive data at the network edge, enabling significant progress in machine learning (ML) applications such as autonomous driving, smart healthcare, and natural language processing~\cite{xie2024efficient,lin2025hierarchical,fang2024automated,zhang2025state,duan2025leed,lin2024efficient,fang2026hfedmoe,sun2025rrto,zhang2024satfed,duan2025llm,fang2024ic3m}.
Conventional centralized learning requires transmitting raw data to edge servers, resulting in prohibitive communication overhead and heightened privacy risks. 
While federated learning (FL)~\cite{song2023distributed,lin2024adaptsfl,hu2024accelerating,hong2026conflict,fang2026aggregation,lin2024fedsn} addresses these issues by enabling collaborative training without data sharing, scaling ML models renders FL impractical for resource-constrained devices. For instance, a large-scale model like OpenELM-3B~\cite{lu2025demystifying} requires approximately 12 GB of storage, resulting in extremely challenging for on-device training on resource-limited edge devices.

To address the limitations of FL, split learning (SL)~\cite{lin2024split,lyu2023optimal,fang2026nsc,lin2025hasfl,tan2026exploiting,lin2025hsplitlora} has emerged as an alternative that offloads computational workloads to edge servers through layer-wise model partitioning, significantly reducing on-device processing requirement.
However, a significant bottleneck in SL arises during training: as the number of devices grows, the massive transmission of smashed data (i.e., activations and gradients) from multiple edge devices to the server becomes an increasingly substantial barrier to deployment.

Recent research has focused on various techniques to compress this smashed data. For example, Zheng~\textit{et al.}~\cite{zheng2023reducing} proposed an activation selection mechanism that retains the top-k highest-magnitude elements, while Eshratifar~\textit{et al.}~\cite{eshratifar2019bottlenet} used an auxiliary model to encode activations into reduced-size representations. Oh~\textit{et al.}~\cite{oh2025communication} introduced a standard deviation-based strategy to discard low-variance components and quantize the remainder.
A key weakness of these existing methods is their reliance on a uniform compression strategy. Smashed data contains diverse types of information (e.g., texture, edges, and object boundaries), each with a varying impact on model performance~\cite{liu20232}. Since all these information types exist within the same feature space, a uniform approach cannot effectively distinguish and process them separately~\cite{gao2020channel}. This often leads to a poor trade-off between communication overhead and model performance, as critical information may be excessively quantized while redundant information is not properly compressed~\cite{niu2025multimodal}.

To address the limitations of existing methods, we propose \underline{SL} with \underline{f}requency-\underline{a}ware \underline{c}ompression (SL-FAC), a communication-efficient SL framework to reduce the transmission overhead of smashed data without sacrificing model performance. SL-FAC consists of two key components: i) adaptive frequency decomposition (AFD) and ii) frequency-based quantization compression (FQC). AFD transforms the smashed data into the frequency domain and decouples it into components corresponding to distinct information types. FQC then customizes the quantization bit widths for each frequency component based on its spectral energy. 
This adaptive approach ensures that critical information is preserved while redundant data is aggressively compressed, optimizing the trade-off between communication overhead and training accuracy.
The main contributions are summarized as follows:
\begin{itemize}
\item To resolve feature-space entanglement, we propose adaptive frequency decomposition, which decouples critical information from noise using spectral decomposition via the discrete cosine transform (DCT)~\cite{yue20213d}.
\item To further optimize compression efficiency, we design frequency-based quantization compression to adaptively customize quantization bit widths for distinct frequency-domain components, thereby achieving efficient compression without compromising training accuracy.
\item We empirically evaluate SL-FAC with extensive experiments, demonstrating that the proposed SL-FAC outperforms state-of-the-art benchmarks by 19.78\% and 6.06\% in test accuracy on the MNIST and HAM10000 datasets, respectively.
\end{itemize}

The rest of the paper is organized as follows. Sec.~\ref{sec:system_design} introduces the system design of SL-FAC. Sec.~\ref{sec:performance_evaluation} provides the performance evaluation. Conclusions are provided in Sec.~\ref{sec:conclusion}.

\section{System Design}
\label{sec:system_design}

\begin{figure}[t]
    \centering
    \includegraphics[width=0.48\textwidth]{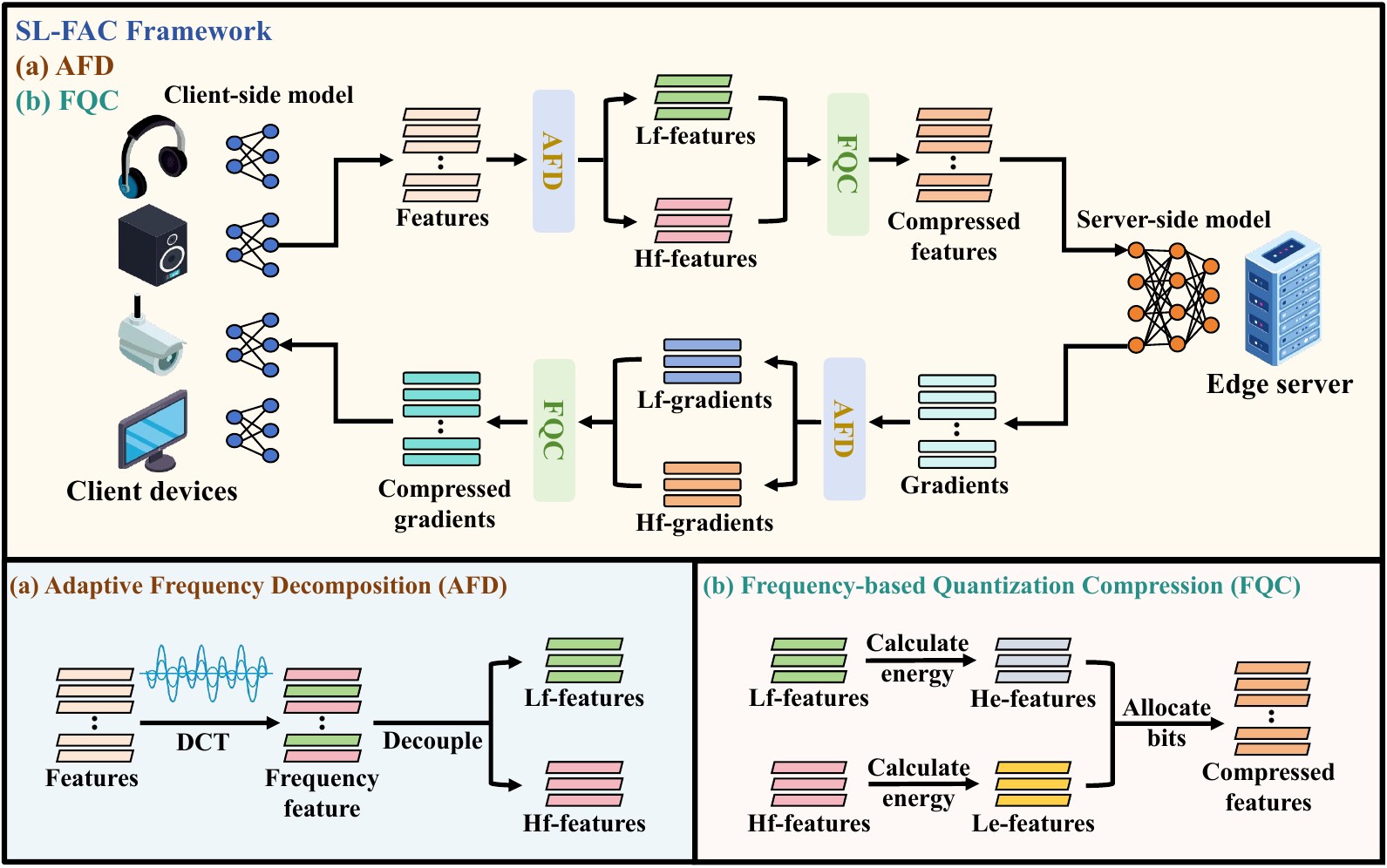} 
    \vspace{-0.5em}
    \caption{The architecture of the proposed SL-FAC framework. (a) AFD for transforming the smashed data into the frequency domain and decoupling it into components corresponding to distinct information types; (b) FQC for customizing quantization bit widths for each component. Here, Lf- and Hf-features (gradients) denote low and high frequency features (gradients), while Le- and He-feature represent low and high energy features, respectively.}
    \label{fig:slfac}
        \vspace{-1em}
\end{figure}

\subsection{Overview}

In this section, we present the architecture and workflow of our proposed SL-FAC framework. 
As shown in Fig.~\ref{fig:slfac}, the global model is split into client-side and server-side sub-models, deployed on the edge device and the edge server, respectively. The training workflow of SL-FAC for one training round follows four steps: 
i)~Each edge device conducts forward propagation of the client-side sub-model on its local dataset to generate activations;
ii)~The activations are processed by AFD to decouple into independent frequency components and then compressed by FQC;
iii)~The compressed activations are transmitted to the edge server to complete the remaining forward and backward passes. The resulting gradients are then processed by AFD and FQC before being sent back;
iv)~Each edge device receives the compressed gradients and updates its client-side sub-model accordingly.

\vspace{-0.5em}

\subsection{Adaptive Frequency Decomposition}
\label{sec:afd}
The massive transmission overhead of smashed data constitutes the primary bottleneck for the large-scale deployment of SL. 
Although existing compression methods attempt to mitigate this issue by filtering out noise, they confront a fundamental limitation: in the spatial domain, critical information and redundant noise are mixed in the same feature space via pixel-level superposition~\cite{gao2020channel,niu2025multimodal}. 
This spatial coupling renders it difficult to isolate noise from useful information.
However, in the frequency domain, this information exhibits completely distinct distribution characteristics. 
The useful information (e.g., image contours and shapes) is highly concentrated in the low-frequency components, whereas noise is isolated in the high-frequency components~\cite{wang2025dctmamba}.
Such disparities make it much easier to identify useful information and filter out noise.

Inspired by this, we propose AFD to decouple these different information types in the frequency domain.
AFD transforms the smashed data into the frequency representation by applying DCT to each channel~\cite{yue20213d}, which is good at representing these variations with a few waveforms. 
We can obtain a set of frequency coefficients $X$, which serve as the weights of the linear combination of cosine basis functions (i.e., frequency components) for information representation. For the $c$-th channel, the coefficient $X_{c,u,v}$ of the $(u,v)$-th frequency component is given by:
\begin{equation}
\begin{split}
X_{c,u,v} = \alpha(u)\beta(v) \sum_{m=1}^{M} \sum_{n=1}^{N} x_{c,m,n} & \cos\left( \frac{\pi}{M} \left( m - \frac{1}{2} \right) (u-1) \right) \\
& \times \cos\left( \frac{\pi}{N} \left( n - \frac{1}{2} \right) (v-1) \right),
\end{split}
\end{equation}
where $x_{c,m,n}$ denotes the element at the $m$-th row and $n$-th column of the $c$-th channel of the smashed data, $M$ and $N$ are the spatial height and width respectively, and $\alpha(u)$ and $\beta(v)$ are the normalization factors to ensure an accurate transformation and prevent errors, which are defined as:
\begin{equation}
\alpha(u) = 
\begin{cases}
\sqrt{\frac{1}{M}}, & u = 1; \\
\sqrt{\frac{2}{M}}, & u \neq 1.
\end{cases}, \quad
\beta(v) = 
\begin{cases}
\sqrt{\frac{1}{N}}, & v = 1; \\
\sqrt{\frac{2}{N}}, & v \neq 1.
\end{cases}
\end{equation}

Once the smashed data are transformed to the frequency
domain, the information richness of the frequency components
can be quantified by the spectral energy. This metric quantifies the maximum amplitude variation of a frequency component: a higher spectral energy indicates a more informative component \cite{ulicny2022harmonic}. Conversely, components with low spectral energy contain less valuable information, such as noise. For the $t$-th training round, the spectral energy $E^t_{c,u,v}$ of the $(u,v)$-th frequency component for the $c$-th channel is simply the square of its coefficient:
\begin{equation}
E^t_{c,u,v} = (X^t_{c,u,v})^2.
\end{equation}

After measuring the information richness of individual frequency components, their contribution to the whole information representation can be further quantified. 
To achieve this, we introduce the cumulative energy ratio, defined as the ratio of the cumulative energy of the first $k$ frequency components (ordered from low to high frequencies via zig-zag scanning) to the total energy of the $c$-th channel. For the $t$-th training round, the cumulative energy ratio $R^t_{c,(k)}$ is expressed as: 
\begin{equation}
R^t_{c,(k)} = \frac{\sum_{i=1}^{k} E^t_{c,(i)}}{\sum_{i=1}^{MN} E^t_{c,(i)}}.
\end{equation}

By accumulating the energy of frequency components, the cumulative energy ratio allows adaptive selection of components that preserve primary information. Specifically, we employ an energy threshold $\theta$ to control the amount of energy retained. Once the ratio $R^t_{c,(k)}$ surpasses $\theta$, the corresponding components of that channel are identified as low-frequency $\mathbf{F}_{l}$, which retains primary information. The rest are treated as high-frequency $\mathbf{F}_{h}$, which carry finer details. This partitioning enables AFD to effectively decouple information types within smashed data.

\vspace{-0.5em}

\subsection{Frequency-based Quantization Compression}
\label{sec:fqc}
Quantization is widely used to encode high-precision data into low-bit representations for communication-efficient training~\cite{fan2024multi,presta2025stanh,ge2024nlic}.
Existing methods \cite{zheng2023reducing,eshratifar2019bottlenet,oh2025communication} typically employ uniform compression for all information in smashed data, overlooking variations in their contribution to model training. This results in excessive quantization of useful information and insufficient compression of less critical elements, eventually degrading overall performance.

To address this problem, we propose FQC, which adaptively allocates quantization bit widths to the low- and high-frequency components $\mathbf{F}_{c,l}$ and $\mathbf{F}_{c,h}$ (see Section~\ref{sec:afd}) based on their spectral energy. 
Specifically, we first compute the average energy of each component for each channel independently.
For the $t$-th round, the average energy $\tilde{E}^t_{c,f}$ of frequency component $f$ in the $c$-th channel is expressed as:
\begin{equation}
\tilde{E}^t_{c,f} = \frac{1}{N_{c,f}} \sum_{k=1}^{N_{c,f}} E^t_{c,f,k},
\end{equation}
where $f \in \{\text{l}, \text{h}\}$ indicates frequency components $\mathbf{F}_{c,l}$ or $\mathbf{F}_{c,h}$,
$N_{c,f}$ represents the number of frequency components in $f$, and $E^t_{c,f,k}$ is the spectral energy of its $k$-th component. 
However, due to the large energy gap between $\mathbf{F}_{c,l}$ and $\mathbf{F}_{c,h}$, direct use of $\tilde{E}^t_{c,f}$ leads to highly unbalanced bit allocation.
Specifically, $\mathbf{F}_{c,l}$ receives a large number of bits, while $\mathbf{F}_{c,h}$ receives only a few bits, making adaptive quantization ineffective.

To address this, we apply a logarithmic transformation to the spectral energy during the allocation of quantization bit widths. The logarithmic transformation compresses the range of high energy values while enhancing distinctions among lower values, thereby reducing polarization in bit assignment. The log-mapped energy for component $f$ of the $c$-th channel is defined as:
\begin{equation}
E^{*,t}_{c,f} = \ln\left(\tilde{E}^t_{c,f} + 1\right).
\end{equation}

The average spectral energy $E^{*,t}_{c,f}$ quantifies the overall information richness of component $f$, where a higher energy indicates that the component contains more information and thus requires more bits to mitigate quantization-induced information loss. Conversely, a lower energy suggests a less informative component, requiring fewer bits to improve compression efficiency. We adaptively allocate quantization bit widths to component $f$ based on $E^{*,t}_{c,f}$ for compression. The allocation is given by:

\RestyleAlgo{ruled}
\LinesNumbered
\begin{algorithm}[t]
\caption{SL-FAC framework.}
\label{alg:slfac}
\setstretch{1}
\scriptsize
\SetKwInOut{Input}{Require}
\SetKwInOut{Output}{Output}
\SetKwProg{Fn}{}{:}{}
\SetKwFunction{afd}{Adaptive frequency decomposition}
\SetKwFunction{fqc}{Frequency-based quantization compression}
\SetKwFunction{fdq}{Quantization decompression}

\Input{$\mathbf{x}$, $\theta$, $b_{\min}, b_{\max}$, $t$.}
\Output{$\tilde{\mathbf{x}}$.}

\Fn{\afd}{
    $C, M, N \gets \text{shape of } \mathbf{x}$\;
    \For{$c = 1, \dots, C$}{
        $\mathbf{X}_{c,:,:} \gets \text{DCT}(\mathbf{x}_{c,:,:})$\;
        \For{$u = 1, \dots, M \text{ and } v = 1, \dots, N$}{
            $E_{c,u,v}^t \gets (X_{c,u,v})^2$\;
        }
        $\{X_{c,(i)}\}_{i=1}^{MN}, \{E_{c,(i)}^t\}_{i=1}^{MN} \gets \text{ZigZagScan}(\mathbf{X}_{c,:,:}, \mathbf{E}_{c,:,:}^t)$\;
        $E_{c,\text{total}}^t \gets \sum_{i=1}^{MN} E_{c,(i)}^t$\;
        \For{$K = 1, \dots, MN$}{
            $R_{c,(K)}^t \gets \frac{\sum_{i=1}^{K} E_{c,(i)}^t}{E_{c,\text{total}}^t}$\;
            \lIf{$R_{c,(K)}^t \ge \theta$}{$k_c^* \gets K$, \textbf{break}}
        }
        $\mathbf{F}_{c,l} \gets \{X_{c,(i)} \mid i \le k_c^*\}$\;
        $\mathbf{F}_{c,h} \gets \{X_{c,(i)} \mid i > k_c^*\}$\;
        $\mathbf{E}_{c,l}^t \gets \{E_{c,(1)}^t, \dots, E_{c,(k_c^*)}^t\}$\;
        $\mathbf{E}_{c,h}^t \gets \{E_{c,(k_c^*+1)}^t, \dots, E_{c,(MN)}^t\}$\;
    }
}

\Fn{\fqc}{
    \For{$c = 1, \dots, C$}{
        \For{$f \in \{l, h\}$}{
            $N_{c,f} \gets |\mathbf{E}_{c,f}^t|$, $\tilde{E}_{c,f}^t \gets \frac{1}{N_{c,f}} \sum_{k=1}^{N_{c,f}} E_{c,f,k}^t$, $E^{*,t}_{c,f} \gets \ln\left(\tilde{E}_{c,f}^t + 1\right)$\;
        }
        $\tau_c \gets \max(E^{*,t}_{c,l}, E^{*,t}_{c,h})$\;
        \For{$f \in \{l, h\}$}{
            $b_{c,f} \gets \left\lfloor b_{\min} + (b_{\max} - b_{\min}) \cdot \phi\left( \frac{E^{*,t}_{c,f}}{\tau_c} \right) \right\rceil$\;
            $x_{c,f,\min} \gets \min(\mathbf{F}_{c,f})$, $x_{c,f,\max} \gets \max(\mathbf{F}_{c,f})$\;
            $\hat{\mathbf{F}}_{c,f} \gets \left\lfloor \frac{\mathbf{F}_{c,f} - x_{c,f,\min}}{x_{c,f,\max} - x_{c,f,\min}} \cdot (2^{b_{c,f}} - 1) \right\rceil$\;
        }
    }
}

\Fn{\fdq}{
    \For{$c = 1, \dots, C$}{
        \For{$f \in \{l, h\}$}{
            $\tilde{\mathbf{F}}_{c,f} \gets \frac{\hat{\mathbf{F}}_{c,f}}{2^{b_{c,f}} - 1} \cdot \left( x_{c,f,\max} - x_{c,f,\min} \right) + x_{c,f,\min}$\;
        }
        $\tilde{\mathbf{X}}_{c,:,:} \gets \text{InverseZigZagScan}(\tilde{\mathbf{F}}_{c,l}, \tilde{\mathbf{F}}_{c,h}, k_c^*)$\;
        $\tilde{\mathbf{x}}_{c,:,:} \gets \text{IDCT}(\tilde{\mathbf{X}}_{c,:,:})$\;
    }
}
\end{algorithm}



\begin{equation}
b_{c,f} = \left\lfloor b_{\min} + (b_{\max} - b_{\min}) \cdot \phi\left( \frac{E^{*,t}_{c,f}}{\tau_c} \right) \right\rceil,
\end{equation}
where $b_{\min}$ and $b_{\max}$ denote the quantization bit width bounds, $\phi(\cdot)$ is the scaling function defined as $\tanh(\frac{\pi}{2} x)$, $\tau_c$ is a dynamic scaling factor set to the maximum spectral energy of the $c$-th channel, i.e., $\tau_c = \max(E^{*,t}_{c,l}, E^{*,t}_{c,h})$, and $\lfloor \cdot \rceil$ is the rounding function.

While DCT coefficients typically follow a Laplace or Generalized Gaussian distribution rather than a uniform one, applying optimal non-uniform quantization incurs prohibitive computational overhead for resource-constrained edge devices. Since the components are grouped into $\mathbf{F}_{c,l}$ and $\mathbf{F}_{c,h}$, these decoupled subsets exhibit consistent spectral energy levels and tightly bounded dynamic ranges. This grouping mechanism effectively mitigates the extreme variance typically observed across the entire frequency spectrum. Consequently, applying min-max linear quantization to these specific subsets serves as a highly computationally efficient and effective approximation. Thus, FQC applies linear quantization using the minimum value $x_{c,f,\min}$ and the maximum value $x_{c,f,\max}$ as boundaries for $\mathbf{F}_{c,l}$ and $\mathbf{F}_{c,h}$, respectively. The quantization procedure is formulated as follows:
\begin{equation}
\label{eq:quan}
\hat{x} = \left\lfloor \frac{x - x_{c,f,\min}}{x_{c,f,\max} - x_{c,f,\min}} \cdot (2^{b_{c,f}} - 1) \right\rceil.
\end{equation}

Once transmitted to the edge server or returned to the edge devices, the compressed smashed data is converted into a floating-point representation before processing by the server-side or client-side sub-model. The process is performed by inverting~(\ref{eq:quan}), as follows:
\begin{equation}
\tilde{x} = \frac{\hat{x}}{2^{b_{c,f}}-1} \cdot \left( x_{c,f,\max} - x_{c,f,\min} \right) + x_{c,f,\min}.
\end{equation}

Finally, the decompressed frequency components are transformed back to the spatial feature space via IDCT for subsequent model processing. The procedure of SL-FAC is summarized in \textbf{Algorithm~\ref{alg:slfac}}.

\section{Performance Evaluation}
\label{sec:performance_evaluation}

\subsection{Experiment Setup}

\subsubsection{Implementation}

Experiments are conducted on a high-performance server equipped with eight NVIDIA RTX 3090 GPUs. To emulate the SL framework, five GPUs are allocated to simulate distributed edge devices, while the remaining GPUs function as the edge server. The software stack includes Python 3.7.2 and PyTorch 1.2.0, which are used to implement the training of image classification tasks.

\subsubsection{Dataset and Model}

We adopt the MNIST~\cite{zhao2025differential} and HAM10000~\cite{tschandl2018ham10000} datasets to evaluate the performance of SL-FAC. Experiments are conducted under both independent and identically distributed (IID) and non-IID settings. In the IID setting, data samples are shuffled and evenly distributed across devices, whereas in the non-IID case, they are partitioned according to a Dirichlet distribution with $\beta = 0.5$. To implement SL-FAC, we employ the ResNet-18 network as the global model, where the first three layers are designed as the client-side sub-model and the remaining layers are assigned as the server-side sub-model.

\begin{figure}[t]
    \vspace{-0.5em}
    \centering
    \includegraphics[width=0.48\linewidth]{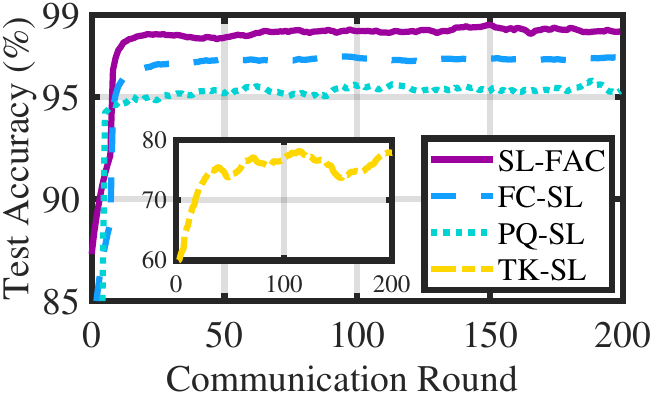}\hfill
    \includegraphics[width=0.48\linewidth]{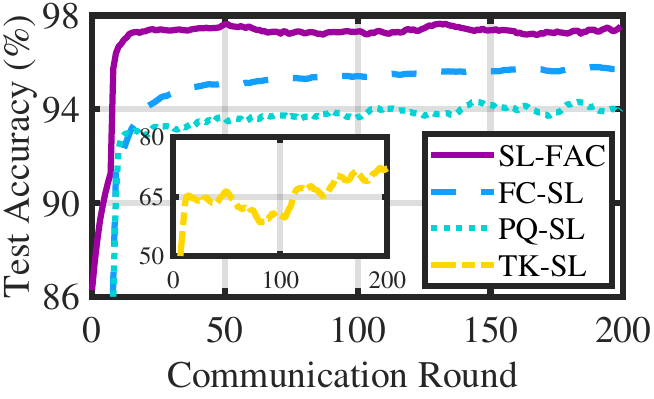}\\[1.0ex]
    \includegraphics[width=0.48\linewidth]{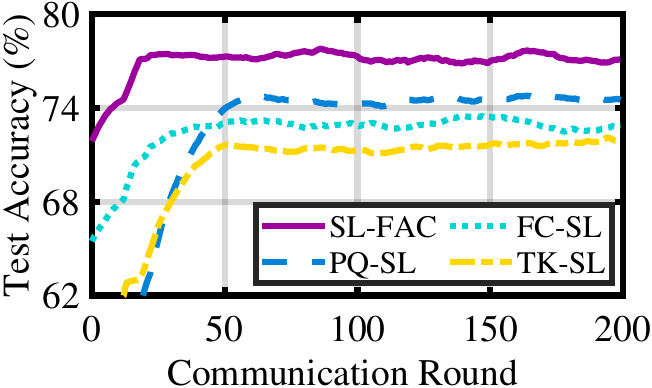}\hfill
    \includegraphics[width=0.48\linewidth]{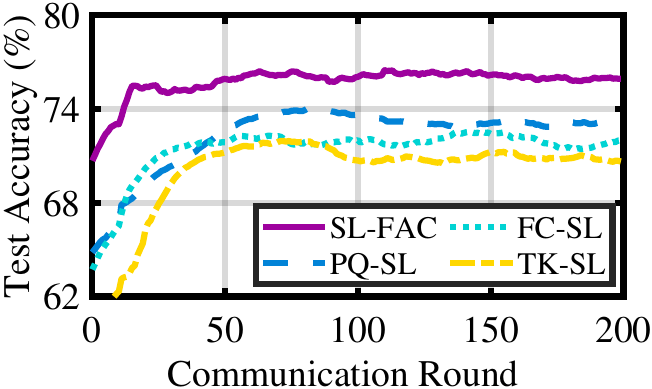}\\[1.0ex]
    \vspace{-1.2em}
    \caption{\rmfamily The training performance of SL-FAC on the MNIST and HAM10000 datasets (from top to bottom), under IID (left) and non-IID (right) settings.}
    \vspace{-1.2em}
    \label{fig:Superiority_SLFAC}
\end{figure}

\subsubsection{Benchmarks}

To evaluate the performance of SL-FAC, we compare it with the following alternatives: i) \textbf{PQ-SL} which is the SL variant of PowerQuant~\cite{yvinec2023powerquant}, which employs a power function to quantize the smashed data for compression; ii) \textbf{TK-SL} which is the SL variant of top-$k$~\cite{zheng2023reducing}, which compresses the smashed data by retaining the top-$k$ largest-magnitude elements and a small subset of the remaining ones; iii) \textbf{FC-SL}~\cite{oh2025communication} is the SL variant of SplitFC~\cite{oh2025communication}, which compresses the smashed data by discarding low-variance features and quantizing the remainder.

\subsubsection{Hyperparameters}


In our experiments, we set the batch size to 128. The quantization bit width is constrained between 2 and 8, with $\theta$ set to 0.9. Unless otherwise stated, the number of edge devices is 5.

\subsection{Superiority of SL-FAC}

As demonstrated in Fig.~\ref{fig:Superiority_SLFAC}, SL-FAC outperforms all baselines across both the MNIST and HAM10000 datasets by achieving the highest convergence accuracy in the fewest communication rounds. Specifically, SL-FAC achieves an accuracy of 98.39\% (97.65\%) on MNIST within only 15 (20) communication rounds, and 77.81\% (76.46\%) on HAM10000 within 30 (40) rounds, under the IID (non-IID) setting. The reason for this is two-fold: On the one hand,  AFD effectively separates informative and redundant components by mapping smashed data into high- and low-frequency components. On the other hand, FQC adaptively allocates bit widths based on spectral energy to preserve critical information while minimizing transmission redundancy, thereby enabling the model to converge with fewer communication rounds. In contrast, FC-SL and TK-SL achieve test accuracies of 95.73\% and 71.56\% under the non-IID setting on MNIST, which are 1.92\% and 26.09\% lower than that of SL-FAC, respectively. The performance gap
is primarily due to their standard deviation- and magnitude-based feature selection strategies, which tend to retain high-magnitude noise while discarding low-magnitude but informative features. 

\vspace{-1em}
\subsection{Impact of Energy Threshold}

\begin{figure}[t]
    \centering
    \vspace{-0.5em}
    \includegraphics[width=0.476\linewidth]{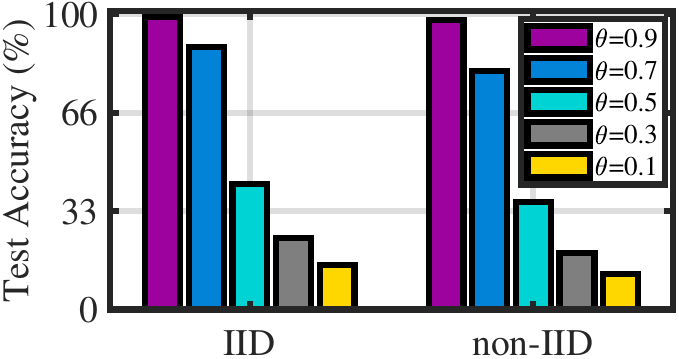}\hfill
    \includegraphics[width=0.476\linewidth]{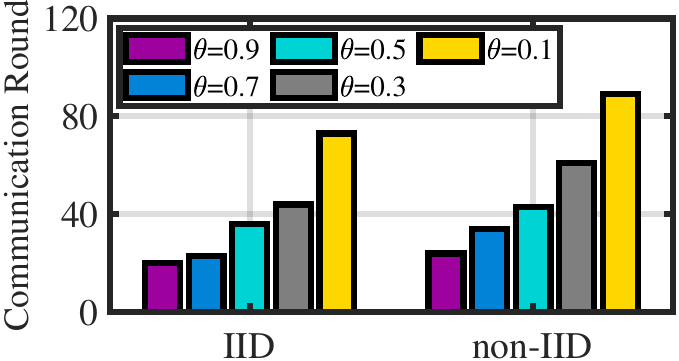}\\[0.5ex]
    \vspace{-1.2em}
    \caption{\rmfamily The impact of energy threshold $\theta$ on model performance using the MNIST dataset under IID and non-IID settings.}
    \vspace{-1.2em}
    \label{fig:Micro_benchmarking1}
\end{figure}

Fig.~\ref{fig:Micro_benchmarking1} illustrates the impact of the energy threshold $\theta$ on model performance on the MNIST dataset under IID and non-IID settings. It can be observed that the performance of SL-FAC improves as $\theta$ increases. The reason is that by increasing $\theta$, more useful information can be filtered by AFD, which allows FQC to allocate more bit widths for its compression, thereby reducing communication overhead without compromising model performance.

\subsection{Ablation Study}

\subsubsection{Adaptive Frequency Decomposition (AFD)}
The first row of Fig.~\ref{fig:ablation_AFD_FQC} illustrates the impact of AFD on training performance by comparing it with magnitude- and STD-based feature selection (i.e., selecting features with the highest magnitude or standard deviation). SL-FAC achieves the highest test accuracy with the fewest communication rounds, outperforming magnitude- and STD-based selection by 1.45\% (1.64\%) and 1.58\% (1.76\%) under the IID (non-IID) setting, respectively. This improvement results from AFD’s use of frequency transformation to effectively filter task-relevant features from smashed data. In contrast, selection based on magnitude or standard deviation introduces noise and fails to prioritize informative features, impairing model convergence and accuracy.

\subsubsection{Frequency-based Quantization Compression (FQC)}

\begin{figure}[t]
    \centering
    \includegraphics[width=0.48\linewidth]{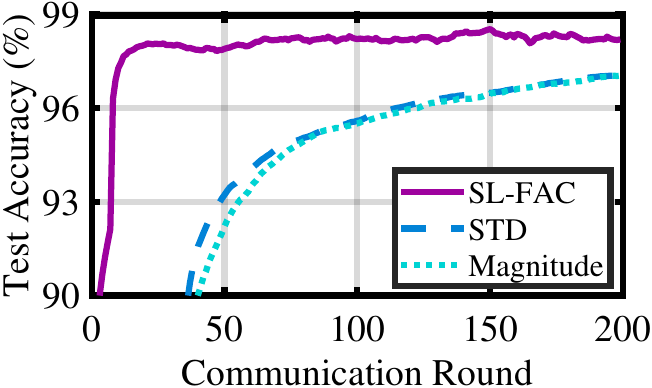}\hfill
    \includegraphics[width=0.48\linewidth]{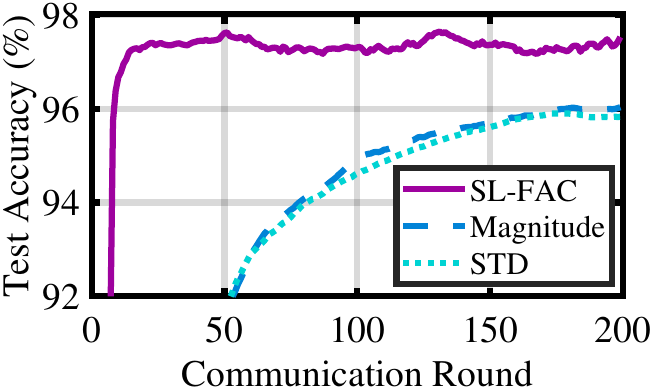}\\[1.0ex]
    \includegraphics[width=0.48\linewidth]{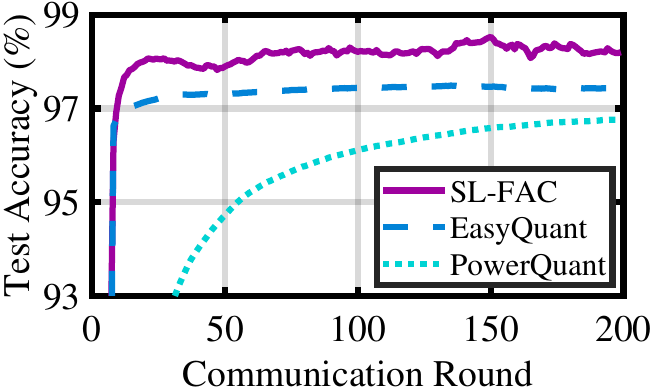}\hfill
    \includegraphics[width=0.48\linewidth]{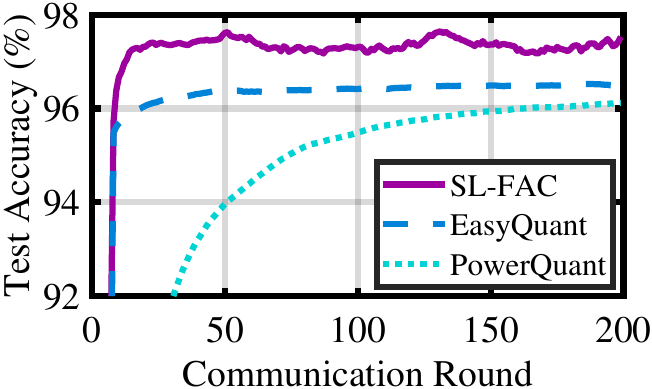}\\[1.0ex]
    \vspace{-1.2em}
    \caption{\rmfamily The ablation experiments for AFD and FQC (from top to bottom) on the MNIST dataset under IID (left) and non-IID (right) settings.}
    \vspace{-1.2em}
    \label{fig:ablation_AFD_FQC}
\end{figure}

The second row of Fig.~\ref{fig:ablation_AFD_FQC} shows the impact of FQC on model performance by comparing it with PowerQuant~\cite{yvinec2023powerquant} and EasyQuant~\cite{tang2023easyquant}. SL-FAC achieves convergence accuracy of 98.39\% (97.65\%), surpassing EasyQuant and PowerQuant by 0.96\% (1.14\%) and 1.65\% (1.52\%) under the IID (non-IID) setting, respectively. While existing benchmarks adopt uniform or power-law scaling, they allocate equal quantization bit widths to both critical and redundant information, resulting in degraded model performance. In contrast, FQC adaptively customizes quantization bit widths based on spectral energy, assigning more bit widths to task-relevant information and fewer to redundant ones, thereby reducing communication overhead without compromising training performance.

\vspace{-1em}
\section{Conclusions}

\label{sec:conclusion}
In this paper, we have proposed SL-FAC, a communication-efficient framework to reduce communication overhead in SL without compromising training performance. SL-FAC addresses a critical limitation of existing methods by moving beyond uniform compression strategies, which consist of two key components: AFD and FQC. AFD transforms smashed data into the frequency domain, effectively decoupling different types of semantic information. FQC then conducts adaptive compression for each component based on spectral energy, ensuring that essential data is preserved while redundant information is aggressively compressed. Extensive experiments have demonstrated that SL-FAC achieves superior performance compared to the state-of-the-art benchmarks, improving test accuracy by 19.78\% and 6.06\% on the MNIST and HAM10000 datasets, respectively.
As a potential future direction, we plan to extend SL-FAC to the training of multimodal large models. 
Given the heterogeneous nature of multimodal data, we aim to leverage AFD and FQC to adaptively compress diverse feature types, thereby reducing communication overhead without compromising cross-modal alignment performance.

{
    \bibliographystyle{IEEEtran}
    \bibliography{references}
}

\end{document}